\documentclass{iosart2c}

\usepackage[T1]{fontenc}
\usepackage{times}%

\usepackage{amsmath}
\usepackage{dcolumn}
\usepackage{graphics}

\usepackage{stfloats}
\usepackage{amsmath, amsfonts}
\usepackage{subfigure}
\usepackage{graphicx}
\usepackage{url}
\usepackage{hyperref}
\usepackage{mathtools}

\newcolumntype{d}[1]{D{.}{.}{#1}}

\firstpage{1} \lastpage{5} \volume{1} \pubyear{2009}

\begin{document}
\begin{frontmatter}                           

%
\title{A real-time multi-constraints obstacle avoidance method using LiDAR} 

\runningtitle{A real-time multi-constraints obstacle avoidance method using LiDAR }

\author[A,B]{\fnms{Wei} \snm{Chen}},
\author[A,B,C]{\fnms{Jian} \snm{Sun}\thanks{Corresponding author. E-mail: sunjian10@mail.xjtu.edu.cn}}
\author[A,B]{\fnms{Weishuo} \snm{Li}}
and
\author[D]{\fnms{Dapeng} \snm{Zhao}}

\runningauthor{Wei Chen et al.}
\address[A]{State Key Laboratory for Strength \& Vibration of Mechanical Structures, School of Aerospace, Xi'an Jiaotong University, Xi'an, PRC, 710049\\
E-mail: cw0523@stu.xjtu.edu.cn}
\address[B]{Shaanxi Engineering Laboratory for Vibration Control of Aerospace Structures, Xi'an Jiaotong University, Xi'an, PRC, 710049}
\address[C]{School of Mechanical Engineering,               Carnegie Mellon University, 5000 Forbes Ave, Pittsburgh, PA 15213, US}
\address[D]{Robotics Institute, School of Computer Science, Carnegie Mellon University, 5000 Forbes Ave, Pittsburgh, PA 15213, US \\
E-mail: sunjian10@mail.xjtu.edu.cn; shuozi2@stu.xjtu.edu.cn; dapengz@andrew.cmu.edu}

\begin{abstract}
Obstacle avoidance is one of the essential and indispensable functions for autonomous mobile robots. Most of the existing solutions are typically based on single condition constraint and cannot incorporate sensor data in a real-time manner, which often fail to respond to unexpected moving obstacles in dynamic unknown environments. In this paper, a novel real-time multi-constraints obstacle avoidance method using Light Detection and Ranging(LiDAR) is proposed, which is able to, based on the latest estimation of the robot pose and environment, find the sub-goal defined by a multi-constraints function within the explored region and plan a corresponding optimal trajectory at each time step iteratively, so that the robot approaches the goal over time. Meanwhile, at each time step, the improved Ant Colony Optimization(ACO) algorithm is also used to re-plan optimal paths from the latest robot pose to the latest defined sub-goal position. While ensuring convergence, planning in this method is done by repeated local optimizations, so that the latest sensor data from LiDAR and derived environment information can be fully utilized at each step until the robot reaches the desired position. This method facilitates real-time performance, also has little requirement on memory space or computational power due to its nature, thus our method has huge potentials to benefit small low-cost autonomous platforms. The method is evaluated against several existing technologies in both simulation and real-world experiments.
\end{abstract}


\end{frontmatter}

\section{Introduction}\label{s1}
In recent years, the autonomous mobile robot technique has made considerable progress and keeps attracting more attention from engineers and researchers around the world. The characteristics of robots, like cost, size, flexibility, safety, etc., drive them to be increasingly popular on various applications from different fields, such as military reconnaissance, surveillance, transportation, traffic monitoring\cite{1,2}. 

One of the major challenges robots having is to avoid obstacles and perform path planning in dynamic environments. Robots should be able to perceive the surrounding environment, be prepared for potential threats, identify new obstacles in the scene, and modify or re-plan trajectory with the latest knowledge, ideally also achieving optimality in some measurable sense at a low cost on memory space and computation.

Generally, path planning can be divided into two categories; one is global path planning that generates an optimal off-line path with prior knowledge of the environment. Even if the environment is pre-mapped, this type of path planning algorithms still suffers from unexpected obstacles that are moving or simply weren’t mapped. The other type does not require the environment to be pre-mapped, but rather, assuming a dynamic environment, work with real-time sensor data to perform online planning locally\cite{3}. For this type of planning, characteristics of sensor data should be taken into consideration during the system design stage to best utilize the information available.

Therefore, the challenges for robot obstacle avoidance algorithms are: (1) insufficient environment information in dynamic environments; (2) sensors' information is often not effectively utilized at each local optimization.

A novel multi-constraints autonomous obstacle avoidance method using LiDAR is proposed in this paper, which enables robots equipped with LiDAR to avoid obstacles autonomously in dynamic situations at a lower cost. It establishes a customized cost function with multi-constraints to analyze effectively the real-time scanning data of the LiDAR, and then extract an optimal sub-goal within the scanning area. Meanwhile, the improved ACO is used to quickly re-plan an optimal sub-path from the current robot position to the sub-goal position. The said process is done iteratively at each step until the pre-specified goal is reached. In our system set up, single-beam 2-dimensional LiDAR, instead of multi-beam 3-dimensional LiDAR, is used because the performance of the former is sufficient for detecting the surroundings and identify obstacles, and it has lower cost and lower requirements on computation and storage. Both simulation and experiments are conducted, to validate our method and demonstrate its effectiveness and availability.

The rest of the paper is organized as follows. Section 2 reviews some important research results related to obstacle avoidance algorithms. Section 3 presents the proposed algorithm. Section 4 demonstrates the simulation and experiment results. Section 5 concludes the paper.

\section{Related work}\label{s2}
Several research studies focus on obstacle avoidance for robot path planning\cite{4}, the methods for solving path planning are as follows: probabilistic, heuristic, and meta-heuristic methods. Probabilistic mainly consists of Rapidly-exploring Random Trees(RRT)\cite{5}, Probabilistic Roadmaps (PRM)\cite{6}. In the heuristic and meta-heuristic approaches are the Artificial Neural Networks(ANN)\cite{7}, Genetic Algorithms(GA)\cite{8}, Simulated Annealing\cite{9}, ACO\cite{10}, Bacterial Foraging Optimization(BFO)\cite{11}. Each of the methods presented above has its strengths and weaknesses. In many situations, some of them are combined to derive the desired path planner in the most effective mode\cite{12,13}. 

The Artificial Potential Field(APF) method was a sophisticated and efficient obstacle avoidance method, first proposed by Oussama Khatib\cite{14} and applied it to obstacle avoidance\cite{11,15,16}, while it is extremely easy to fall into the local minima. Bence Kovács\cite{17} presented a method from animal motion attributes based on APF for robot path planning, and the Bug algorithm was used to solve the local minima problem of APF. However, it has not solved the disadvantages that the Bug algorithm is incapable of exploiting the sensor’s data and still cannot ensure that the path is optimal. Dieter Fox proposed a Dynamic Window Approach(DWA) in\cite{18,19}, which took into account the inertia factor of the robot and is suitable for robots with high speed and robots with limited motor torque; but, it cannot be used in unstructured dynamic environments. In\cite{11} BFO algorithm produced impressive simulation results for obstacle avoidance on a mobile robot path planning, this method requires many samples for estimation, the more samples, the greater the memory consumption, the higher the complexity of the algorithm\cite{20}; conversely, the fewer the particles, the worse the path smoothness. Reference\cite{21} proposed an obstacle avoidance bubble bug algorithm (BBA), it defined a bubble around the robot, the radius of the bubble indicates the range of the sensor, the strategy was to limit the robot to the maximum distance from the obstacle all the time. Some scholars are working to solve the problem of robot navigation and obstacle avoidance by Simultaneous Localization and Mapping (SLAM)\cite{22,23} based on LiDAR or vision\cite{24,25,26,27}. 

In contrast to the mentioned approaches, this paper focuses on building a multi-constraints mathematical model to best utilize the real-time scanning information of LiDAR in a dynamic environment. As a result, the robot is able to, in a real-time manner, computes the optimal path to avoid unexpected threats and uncertain obstacles at a lower cost.

\section{Proposed method}\label{s3}
In this section, the cost function with multi-constraints is introduced to analyze the local known information, including the current position and orientation (together, the pose) of the robot, the LiDAR scanning area (obstacle areas and collision-free space), and the goal position.
\subsection{Environment model}\label{s3.1}
LiDAR is mounted on the top of the robot and used to detect environment information in a 2D scanning plane, with a maximum scanning range. Note that the 2D LiDAR can only scan one plane. In practice, if the height of the obstacle is less than the height of the LiDAR, it cannot be detected. Therefore, the installation height of LiDAR needs to weigh the actual application environment and should not be fixed. At each scanning period, LiDAR gives 360-degree surrounding scanning and generates $N_{p}$ data points, also called lase-point cloud. Scanning data can be expressed as $\left(d_{i}^{t}, \theta_{i}^{t}\right)$, where the $d_{i}^{t}$ and $\theta_{i}^{t}$ respectively denotes the relative distance and orientation centered on the robot, with the LiDAR scanned objects for $i=1, \ldots, N_{p}$ at the time $t$. As shown in figure 1, the original data of the LiDAR in polar coordinates is shown, where the green dots indicate the obstacles, $R$ represents the LiDAR scanning radius, and the  white triangle denotes the robot.

\begin{figure}[htbp]
\vspace{-0.2cm} 
\setlength{\abovecaptionskip}{0.1cm}  
\includegraphics[scale = 0.35]{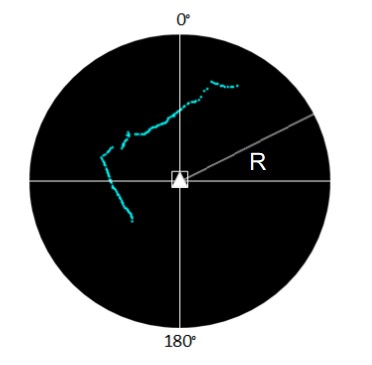}
\caption{Original data of LiDAR}\label{f1}
\end{figure}

Within the scanning area, every object can be simplified as multiple points. If there are obstacles, a series of point sets are generated to represent obstacles; otherwise, there are no data points. It is not convenient to use the original LiDAR data directly to represent surroundings because of the data points are discrete. A single point does not indicate an obstacle, but clusters the data points. There have been several studies focusing on clustering methods for point cloud data. In this paper, the scanning area is equally divided into $N_{m}$ sectors, and each sector is given a unique number $i d\left(i d=1, \ldots, N_{m}\right)$. Generally, that is meaningful to equally divide the scanning area, the smaller the sector division, the more flexible the path planning, while the amount of calculation is increased. All data points will be distributed in the sector corresponding to the $\left(d_{i}^{t}, \theta_{i}^{t}\right)$ within the scanning area. In the way, only need to pay attention to which sectors are distributed with data points; the remaining sectors belong to collision-free space. A local grid map is constructed within the LiDAR scanning area, which is used to analyze the current environment.

The scanning area is further divided into small grid cells that can accurately locate the obstacle in the specific grid units. Because roughly creating a grid map in a circular scanning area of radius $R$ may produce some incomplete grid units, making it impossible to assign a corresponding coordinate position to some grid units. In order to solve the problem that the incomplete grid cells cannot be assigned coordinates, in this paper, the largest square space in a circular scanning area is used to replace the LiDAR scanning area, and then the square is divided into several grid cells. So each clustered data points can be allocated in a specific grid.

In this paper, the robot moves on the horizontal ground and obtains real-time pose information by reading the odometry. Its position and orientation can be simplified as $P_{robot}^{t}\left(x_{r}, y_{r}, \psi_{r}\right)$ in the 2D space at time $t$, and $\psi_{r}$ indicates the yaw angle. In fact, given the error of the odometry, the application environment is limited to small indoor scenes. It can be easily calculated that  grid units position $P_{grid}^{t, i}\left(x_{i}, y_{i}\right)$ based on the $P_{robot}^{t}\left(x_{r}, y_{r}, \psi_{r}\right)$ at time $t$. 
\setlength{\abovedisplayskip}{3pt}  
\setlength{\belowdisplayskip}{3pt}
\begin{equation} 
\left(\begin{array}{l}
x_{i} \\
y_{i}
\end{array}\right)=\left(\begin{array}{l}
x_{r} \\
y_{r}
\end{array}\right)+\left(\begin{array}{l}
d_{i}^{t} \cdot \cos \left(\psi_{r}-\theta_{i}^{t}\right) \\
d_{i}^{t} \cdot \sin \left(\psi_{r}-\theta_{i}^{t}\right)
\end{array}\right)
\end{equation}

In the grid map, an expansion coefficient is set for each obstacle, and if a grid unit is occupied by an obstacle, its adjacent eight grid units are not included in the collision-free space. However, how to logically re-plan an optimal sub-path among these collision-free grids to achieve the avoidance obstacles at a lower cost is the problem to be solved in this paper.

\subsection{The cost function}\label{s3.2}
The robot can re-plan an optimal path online toward sub-goal and successfully avoid unexpected obstacles or threats at a lower cost. As a sub-goal selected at a specific moment, it employs the key criterion for convergence to the goal position. On the other hand, the sub-goal is locally optimal and globally convergent, because the cost function based on multi-constraints is capable of analyzing thoroughly the detection information of the LiDAR. Concerning known local environment information, each grid cell in collision-free space is estimated by the cost function, the result of the cost function denotes the cost of the corresponding grid cell as a sub-goal, the smaller the value, the smaller the cost.

The significance of the customized cost function consisting of $\!d s\left(P_{grid}^{t, j}, P_{goal}\right)\!, \!\theta_{1}\left(P_{{robot}}^{t}, P_{grid}^{t, j}\right)\!$ and $\!\theta_{2}\left(P_{grid}^{t, j}, P_{{goal }}\right)\!$ is to comprehensively evaluate the real-time information, including the current pose information of the robot, the LiDAR scanning area, and the goal position, to generate the sub-goal. $P_{goal}$ is the goal position information, $P_{robot}^{t}$ denotes the robot pose information at the time $t,$ and $P_{{grid }}^{t, j}$ indicates the position information of the $j$-$th$ grid unit in collision-free space at the time $t$.

$\!ds(P_{grid}^{t, j}, P_{goal})\!$ is used to calculate the euclidean distance between $P_{{grid }}^{t, j}$ and $P_{{goal}}$ at time $t$, an essential contribution $\!ds(P_{grid}^{t, j}, P_{goal})\!$ is to ensure the global convergence of sub-goal. In the specific implementation, the scanning information is maximized to achieve the local optimal obstacle avoidance result. Only the most marginal grid units in the collision-free space are selected, which can promote the processing efficiency. The higher the distance to the obstacle, the more flexible the path to avoid the obstacle, which means a more substantial area around the robots are unobstructed.

$\!\theta_{1}(P_{robot}^{t}, P_{grid}^{t, j})\!$ returns the azimuth angle from $P_{robot}^{t}$ to $P_{{grid }}^{t, j}$ at time $t$, which is angle with the yaw$(\psi)$ of robots. As one of the constraint factors, $\!\theta_{1}(P_{robot}^{t}, P_{{grid }}^{t, j})\!$ embodies the angle deviation relationship between the robot and the grid cells in collision-free space.

$\!\theta_{2}(P_{grid}^{t, j}, P_{{goal}})\!$ generates the azimuth angle from $P_{{grid }}^{t, j}$ to $P_{{goal}}$ at time $t$, which is used to indicate the magnitude of deviation of the selected grid unit from the goal position. The additional contribution of constraints factors $\!\theta_{1}(P_{robot}^{t}, P_{{grid}}^{t, j})\!$ and $\!\theta_{2}(P_{grid}^{t, j}, P_{{goal}})\!$ are to ensure the smoothness of the path.

The proposed cost function is defined as Eq.(2):
\begin{equation}
F(j)=\left|\begin{array}{ll}
\beta \\
\alpha \\
\omega
\end{array}\right|^{T} \cdot\left|\begin{array}{l}
\theta_{1}\left(P_{robot}^{t}, P_{grid}^{t, j}\right) \\
d s\left(P_{grid}^{t, j}, P_{goal}\right) \\
\theta_{2}\left(P_{robot}^{t}, P_{grid}^{t, j}\right)
\end{array}\right|
\end{equation}
$\alpha, \beta$ and $\omega$ is the weight coefficient respectively.

The above three constraints are not directly added after calculation, but are normalized first, and then added. Since the cost function is composed of multiple constraints, and different constraints represent information in different dimensions, such as distance and different angle. To prevent a certain constraint from being too dominant in the cost function,and the normalized method is used to smooth the constraint in this paper.
\begin{equation}
\!\left\{\begin{aligned} \operatorname{nor}\left(ds(j)\right)=& \frac{ds\left(P_{grid}^{t, j}, P_{goal}\right)}{\sum\limits_{j} ds\left(P_{grid}^{t, j}, P_{goal}\right)} 
  \\ \operatorname{nor} \left(\theta_{1}(j)\right)=& \frac{\theta_{1}\left(P_{robot}^{t}, P_{grid}^{t, j}\right)}{\sum\limits_{j} \theta_{1}\left(P_{robo t}^{t}, P_{grid}^{t, j}\right)} 
  \\ \operatorname{nor} \left(\theta_{2}(j)\right)=& \frac{\theta_{2}\left(P_{grid}^{t, j}, P_{goa l}\right)}{\sum\limits_{j} \theta_{2}\left(P_{grid}^{t, j}, P_{goal}\right)} \end{aligned}\right.\!
\end{equation}
Since the latest local known information is different each time, the number of available grids is different each time.

Finally, with the appropriate weight parameters, the normalized constraints are introduced into the cost function to estimate the cost for each available grid cell and choose the minimum cost.
\begin{equation}
\begin{split}
P_{s g}(x, y)=\min \left(\left|\begin{array}{l}
\beta \\
\alpha \\
\omega
\end{array}\right|^{T} \cdot\left|\begin{array}{l}
\operatorname{nor}\left(\theta_{1}(j)\right) \\
\operatorname{nor}(d s(j)) \\
\operatorname{nor}\left(\theta_{2}(j)\right)
\end{array}\right|\right) 
\end{split}
\end{equation}

Based on the latest local known information each time, the grid cell corresponding to the minimum cost value is selected as the sub-goal $P_{s g}(x,y)$, which satisfies the optimization criterion such as distance, time, and cost in a limited scanning area, so the robot is capable of immediately responding to any moving obstacles within the scanning area of LiDAR. In the grid map of the scanning area, the path of the robot is connected by multiple grid cells. Although the robot is capable of moving in any direction in a 2D plane, the motion model of the robot is further defined in this paper, assuming that the robot is only allowed to move along the front, rear, left, right, and diagonal lines between adjacent grids. The resolution of the grid cell is small enough not to affect the smoothness of the path. Therefore, how to quickly re-plan a sub-path from the current robot position to the sub-goal position is urgently needed. Generally, a straight motion is considered as the primary solution because straight-line motion usually indicates that the path is the shortest, while the shortest path is not necessarily the optimal path in many practical problems.

\subsection{Improved ACO algorithm}\label{s3.3}

The improved ACO, in this paper, is used to plan optimal sub-path from the latest robot position to the latest defined sub-goal, The robot moves one step along the sub-path, then sets the current position as the start point, extracts the local sub-goal position based on the latest local known information, and calls ACO to re-plan the sub-path to avoid moving obstacles until it reaches the goal position. The flowchart of the proposed method is shown in figure $2$. 
\begin{figure}[ht] 
\vspace{-0.2cm} 
\setlength{\abovecaptionskip}{0.1cm}  
\includegraphics[scale = 0.35]{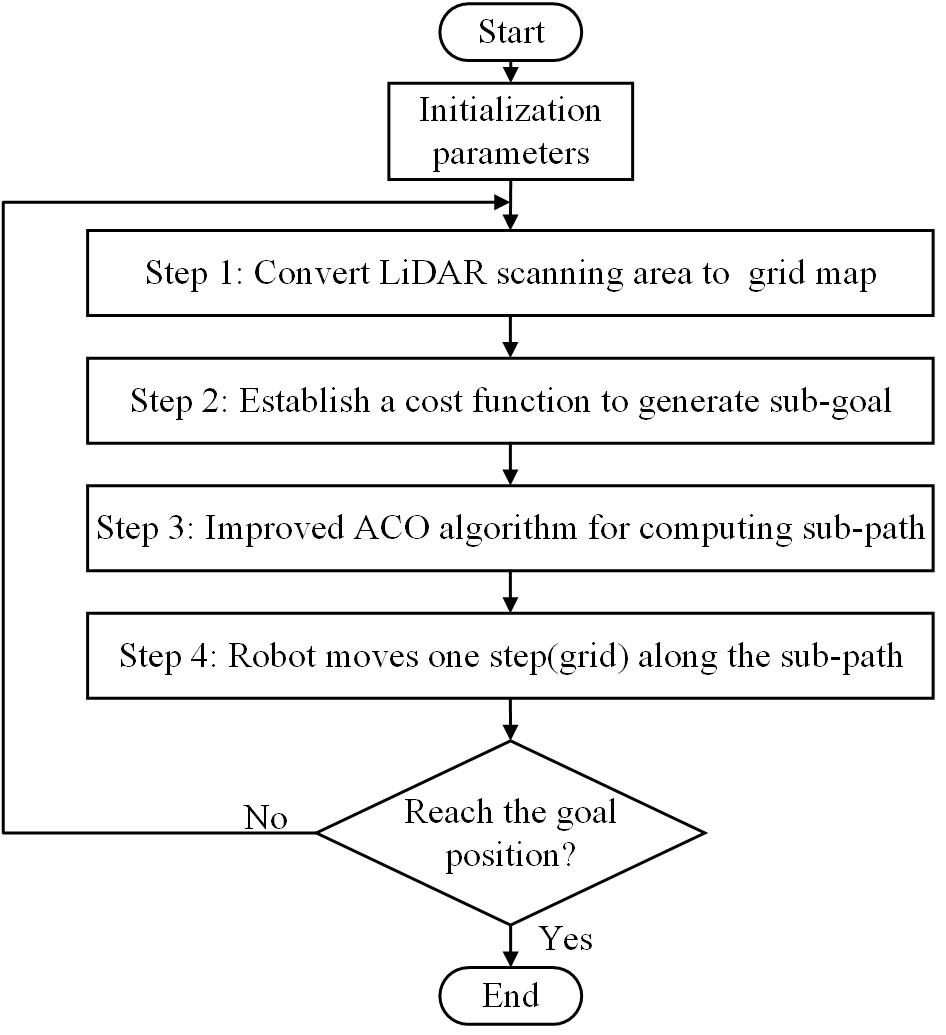}
\caption{ Flowchart of the proposed algorithm}\label{f3}
\end{figure}

Note that the next sub-goal position is calculated if and only if the robot reaches the sub-goal point, which will cause a delay in updating the environmental information, and the robot will not be able to respond to moving obstacles before reaching the sub-goal position. 

The ACO algorithm has the characteristics of global optimization, parallelism, and strong robustness, and has been widely used in offline path planning problems. In this paper, the ACO algorithm is used to online re-plan the optimal sub-path of the robot from the current position to the sub-goal position in the grid map of the scanning area. The fundamental definition of the ACO algorithm is typically given by Eq.(5):
\begin{equation} 
\!P_{ij}^{k}(t)\!=\left\{\begin{array}{ll}
\frac{\!\left[\tau_{i j}(t)\right]^{\phi}\! \cdot \!\left[\eta_{i j}(t)\right]^{\gamma}\!} {\!\sum\limits_{s \in J_{k}(i)} \left[\tau_{i s}(t)\right]^{\phi} \cdot \left[\eta_{i s}(t)\right]^{\gamma}\!}, &  \text {if} j \in \!J_{k}(i)\! \\
0, & \text {if} j \notin \!J_{k}(i)\!
\end{array}\right.
\end{equation}

where the ${P}_{i j}^{k}(t)$ is the transition probability of ant $k$ moving from $i(x_{i}, y_{i})$ to $j(x_{j}, y_{j})$ at the time $t ; \tau_{i j}(t)$ denotes the pheromone from $i(x_{i}, y_{i})$ to $j(x_{j}, y_{j})$ at the time $t$; $\phi$ is the relative importance of pheromone, $\gamma$ is the relative importance of heuristic factor, $J_{k}(i)$ is a collection of positions that allows the ant to walk next; and $\eta_{i j}(t)$ represent the heuristic factor from $i(x_{i}, y_{i})$ to $j(x_{j}, y_{j})$ at the time $t$,

\begin{equation} 
\eta_{i j}(t)=\frac{1}{d_{i j}}
\end{equation}
$d_{i j}$ is the distance from $i(x_{i}, y_{i})$ to $j(x_{j}, y_{j})$.

After all ants complete once iteration, the pheromone update method is as follows:
\begin{equation} 
\tau_{i j}(t+1)=(1-\rho) \tau_{i j}(t)+\Delta \tau_{i j}(t)
\label{eq5}
\end{equation}
\begin{equation} 
\Delta \tau_{i j}(t)=\sum_{k=1}^{m} \Delta \tau_{i j}^{k}(t)
\end{equation}
\begin{equation} 
\Delta \tau_{i j}^{k}(t)=\left\{\begin{array}{ll}{\frac{Q}{L_{k}},} & {\text { if ant } k \text { travels on edge }(i, j)} \\ {0,} & {\text { otherwise }}\end{array}\right.
\end{equation}
where the $\rho \in(0,1)$ is the rate of pheromone evaporation, and $1-\rho$ is the residual
pheromone factor; $\Delta \tau_{i j}(t)$ is the amount of deposited pheromone from $i\left(x_{i}, y_{i}\right)$ to $j\left(x_{j}, y_{j}\right)$ at time $t$, and $\Delta \tau_{i j}^{k}(t)$ is the amount of deposited pheromone from $i\left(x_{i}, y_{i}\right)$ to $j\left(x_{j}, y_{j}\right)$ by ant $k$ at time $t$; $Q$ is positive constant; $\mathbf{m}$ and $N$ are the set of ants colony and the number of iterations, respectively; $L_{k}$ indicates the path length by ant $k$ in this iteration.

Based on the traditional ACO algorithm, to enhance the stability of robot movement, corner heuristic information is introduced into the path transition probability function to reduce the number of large corners and the number of corners in the path. The improved path transition probability function is as follows:
\begin{equation} 
\!P_{i j}^{k}(t)\!\!=\left\{\!\begin{array}{ll}
\frac{\!\left[\tau_{i j}(t)\right]^{\phi}\! \cdot \!\left[\eta_{i j}(t)\right]^{\gamma}\! \cdot \!\left[v_{i j}(t)\right]\!}{\sum\limits_{s \in \!J_{k}(i)\!} \!\left[\tau_{i s}(t)\right]^{\phi}\! \cdot \!\left[\eta_{i s}(t)\right]^{\gamma}\! \cdot \!\left[v_{i s}(t)\right]\!}, & \!\text { if }\! \!j\! \!\in\! \!J_{k}(i)\! \\
0, & \!\text { if }\! \!j\! \!\notin J_{k}(i)\!
\end{array}\right.
\end{equation}
$v_{i j}(t)$ is the corner heuristic function from $i\left(x_{i}, y_{i}\right)$ to $j\left(x_{j}, y_{j}\right)$ at time t. $v_{i j}(t)$ and
$\theta_{i j}$ are inversely proportional.

\begin{equation} 
v_{i j}(t)=\left\{\begin{array}{l}
\frac{1}{\theta_{i j}}, \text { if } \theta_{i j} \neq 0 \\
1, \quad \text { othersize }\end{array}\right.
\end{equation}
\begin{equation} 
\theta_{i j}=\operatorname{atan} 2\left(y_{j}-y_{i},x_{j}-x_{i}\right)
\end{equation}

In the pheromone update phase, on the one hand, the traditional ACO algorithm updates the pheromone of all ants, including the worst ants, which will be misleading for the next generation. On the other hand, only the length of the path is considered, as shown in Eq. (7), (8) and (9). To coordinate the smoothness of the path and the convergence speed of our method. each ant gets a score based on its path length and number of corners, and ranks the score in descending order. The improved pheromone update rule is that the pheromone increment is determined by weighting the path length and the number of path corners, which can ensure the smoothness of the sub-path. At the same time, only some good ants with higher scores are updated to avoid the impact of the worst ants on the offspring ants, thereby improving the convergence speed of the algorithm.
\begin{equation} 
\Delta \tau_{i j}^{k}(t)=\left\{\begin{array}{cc}
\frac{Q}{\mathrm{S}\left(w_{k}\right)}, & \operatorname{rank} \leq(m-1) \\
0, & \text { otherwise }
\end{array}\right.
\end{equation}
\begin{equation} 
w_{k}=|\delta \quad \zeta| \cdot\left|\begin{array}{l}
L_{k} \\
T_{k}
\end{array}\right|, \quad(k=1, \ldots, m)
\end{equation}

$w_{k}$ is the weighted score of the path length and number of corners of ant $k$, $\mathrm{S}\left(w_{k}\right)$ is the sort function to sort $w_{k}$ from large to small, $\delta$ and $\zeta$ are the positive weight coefficients, $T_{k}$ indicates the number of corners of the path generated by ant $k$. $\mathrm{rank}$ is the ant order after reordering

The implementation steps of the improved ACO method are as follows:

Step 1: Algorithm initialization, $P_{robot}^{t}$ and $P_{sg}(x,y)$ are initialized to the ant colony start position and sub-goal position respectively, and the search space is the known grid map of the scanning area.

Step 2: Calculate the path transition probability of each ant according to the $\operatorname{Eq}(10)$ and choose the next feasible position $j$ according to the roulette rule. If $j$ is the sub-goal point, the step 3 is performed; otherwise, still, execute the step 2.

Step 3: After all ants have completed an iteration, $H_{n}^{k}$ records the path trajectory generated by ant $k$ in the $n$-$th$ iteration. Find the ant in $H_{n}^{k}$ that does not reach the sub-goal point as the set $\mathbf{U}$. Randomly select an ant $s$ to re-modify its path trajectory in this iteration.
\begin{equation} 
\begin{split}
s=\left\{\begin{array}{c}{\operatorname{random}(\mathbf{U}), \text { if } \mathbf{U} \neq { null }} \\ {\operatorname{random}(\mathbf{m}), \text { if } \mathbf{U}= { null }}\end{array}\right.
\end{split}
\end{equation}
\begin{equation} 
H_{n}^{s}=f\left(H_{n-1}^{k}\right), \quad(k=1, \ldots, m)
\end{equation}
$f\left(H_{n-1}^{k}\right)$ is the optimal path before the $n$-$th$ iteration. When set $U$ is valid, randomly select one ant $s$ in $U$ and assign $f\left(H_{n-1}^{k}\right)$ to it ; if all ants reach the target point, randomly select anyone to assign.

Step 4: According to the $\operatorname{Eq}(14)$ to calculate the score of each ant, and then update the pheromone of the selected ant according to the  $\operatorname{Eq}(13)$.

Step 5: Determine whether the number of iterations reaches the maximum number of iterations $N$ . If it does, the step 6 is performed. Otherwise, steps 2 to 5 are executed cyclically.

Step 6: The optimal sub-path described by $\operatorname{Eq}(17)$, $l_{1}$ represents the position of the grid cell corresponding to $P_{robot}^{t}$, $l_{\mu}$ is the position of the grid cell corresponding to sub-goal $P_{sg}(x,y)$.
\begin{equation} 
P_{path}\left(l_{1}, l_{2}, \ldots, l_{\mu}\right)=f\left(H_{N}^{k}\right)
\end{equation}
adding $P_{path}\left(l_{1}, l_{2}, \ldots, l_{\mu}\right)$ to the desired path, the robot moves one step along the sub-path to the next position $l_{2}$. Repeat the steps 1 to 6 until the robot reaches the goal position.

The convergence speed of improved ACO algorithm is shown in figure $3$. The black solid line shows the convergence speed of the improved ACO algorithm. The concept of corners is involved in both the transition probability phase and the pheromone update phase to build up the smoothness of the path. Update only some good ants to avoid the impact of the worst ants on the offspring ants. It can be seen that the improved ACO has faster convergence speed. At the same time, it should be pointed out that it is not advisable to assign multiple ants in once iteration, which will increase the probability that the algorithm falls into the local minima.
\begin{figure}[hbp] 
\vspace{-0.2cm} 
\setlength{\abovecaptionskip}{0.1cm}  
\includegraphics[scale = 0.4]{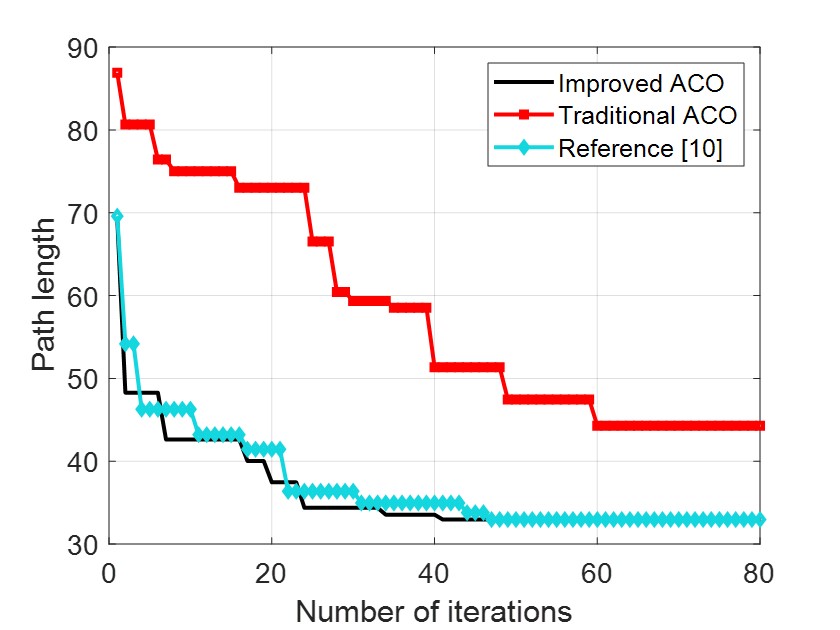}
\caption{Comparison of ACO convergence}\label{f2}
\end{figure}

\section{Experimental results}\label{s4}
\subsection{Simulation results}\label{s4.1}
Before the real experiment of the robot, a simulation of the algorithm is implemented in MATLAB to compare the effect of different parameters on the results. In order to simulate the actual environment more realistically, an obstacle map is loaded by a grid map in the simulation program, which corresponds to a 40m square real area. Black areas indicate the collision-free area, and the white grids denote the obstacles, which also called occupied units. In the simulation of this paper, each grid cell in the grid map corresponds to a 1.5m square area. Eight grid cells adjacent to the occupied unit are excluded from collision-free space as a safe distance. 

\subsubsection{Simulation 1 - environment test}\label{s4.1.1}
In order to demonstrate the availability and optimality of the proposed algorithm under multiple obstacles situation, figure 4 shows the simulation results of the proposed algorithm in more detail in the form of multiple consecutive instants. 
\begin{figure*}[htbp]
\vspace{-0.1cm} 
\setlength{\abovecaptionskip}{-0.1cm}
\centering
\subfigure[step = 1]{
\includegraphics[scale = 0.19]{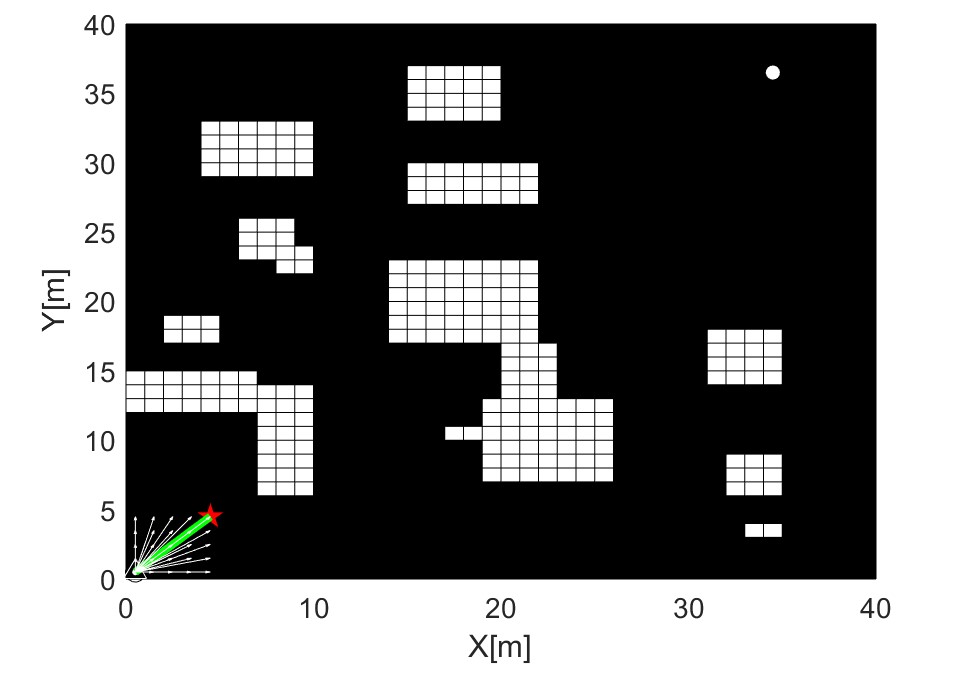}
}\hspace{-10mm}
\subfigure[step = 20]{
\includegraphics[scale = 0.19]{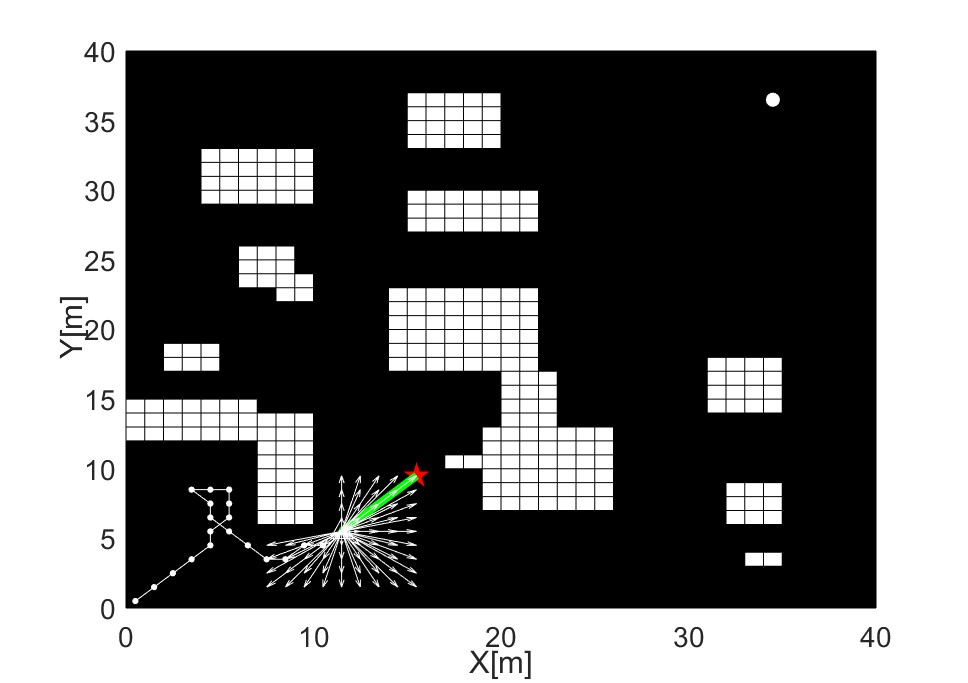}
}\hspace{-10mm}  
\subfigure[step = 30]{
\includegraphics[scale = 0.19]{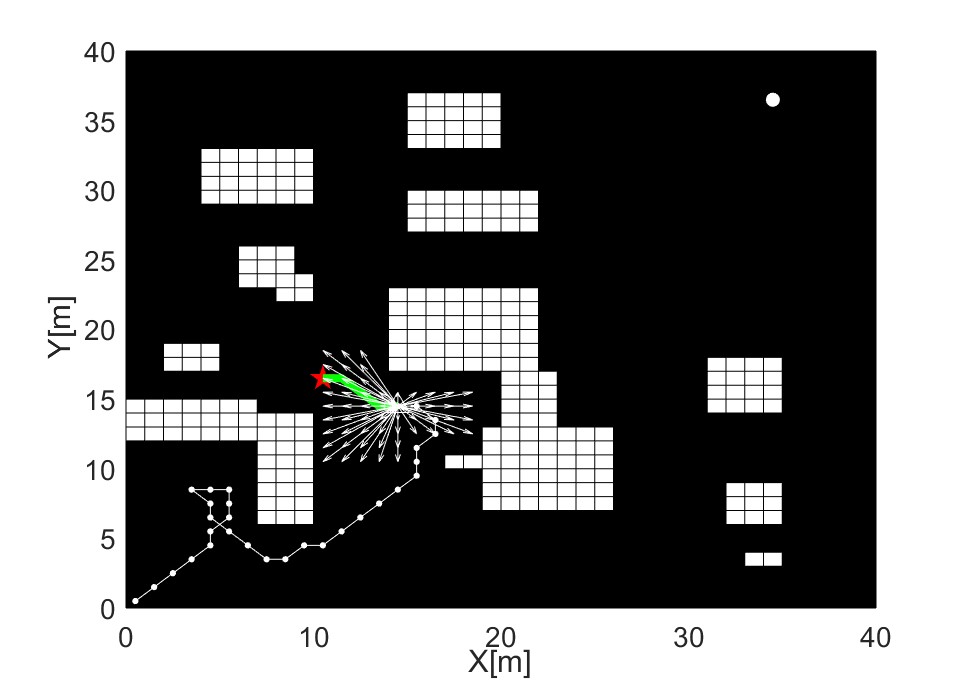}
}\vspace{-4mm}  
\subfigure[step = 35]{
\includegraphics[scale = 0.19]{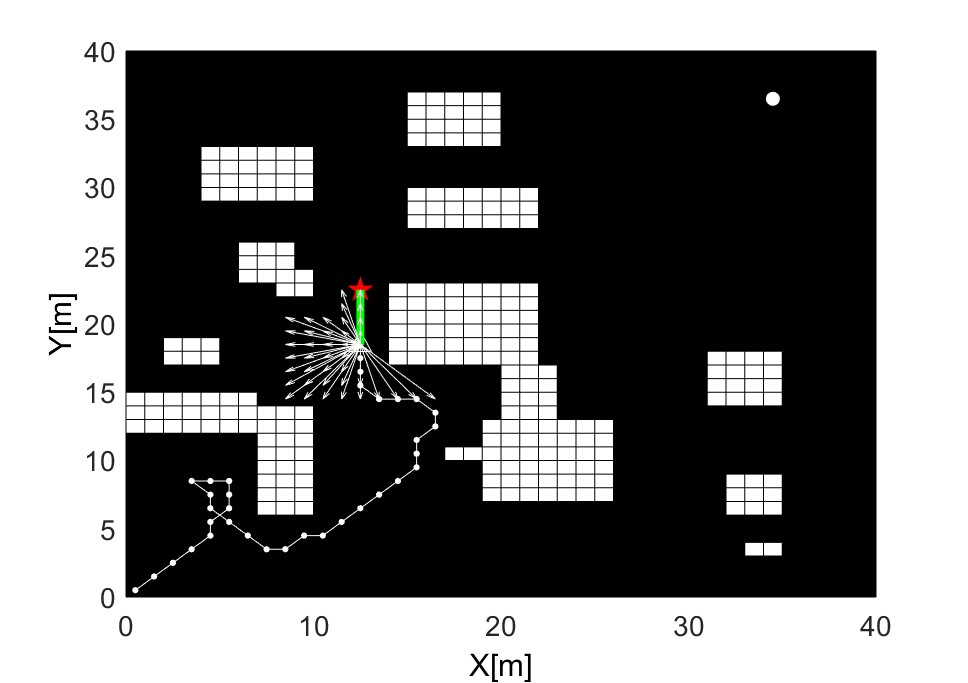}
}\hspace{-10mm}
\subfigure[step = 40]{
\includegraphics[scale = 0.19]{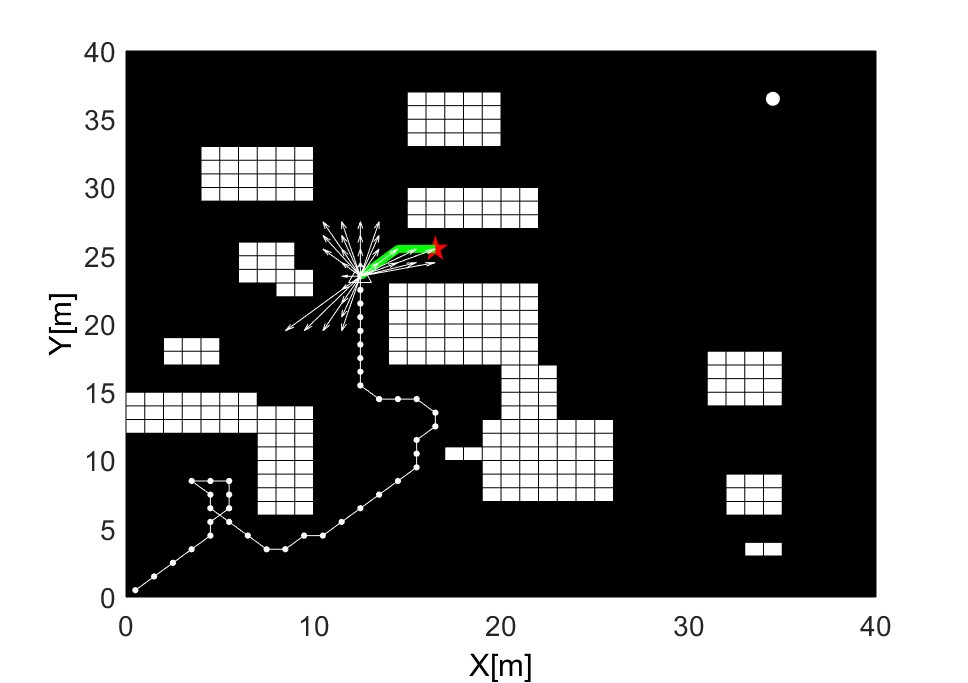}
}\hspace{-10mm}
\subfigure[end]{
\includegraphics[scale = 0.19]{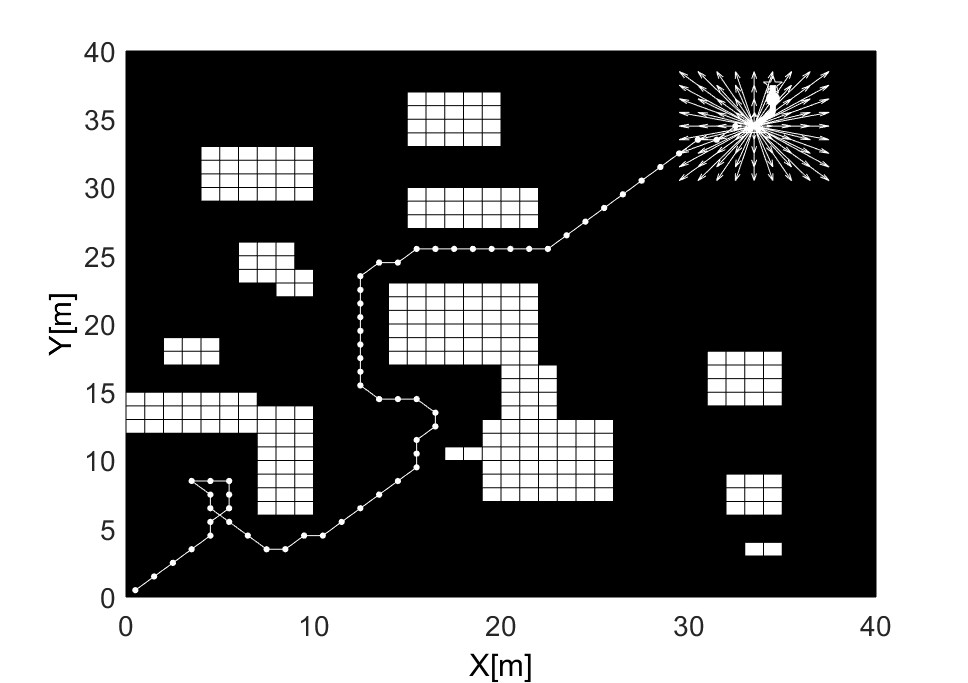}
}
\caption{ Consecutive instants in the map }
\end{figure*}

In real application, the robot only knows the information within the LiDAR detection range at a certain moment. The detection radius of the LiDAR is four grid cells centered on the robot, which includes a total of 80 grid cells adjacent to each other. The simulation program window shows that the white vector arrows denote the LiDAR scanning area, and the solid white circle is the goal position. The red pentagram marks the sub-goal position calculated by the cost function proposed in this paper at a specific moment, and the solid green line is the corresponding sub-path that re-planed by the improved ACO algorithm in real-time. Figure 5 shows that the robot adjusts the pose in real-time during the movement to meet the current state based on the latest local information.
\begin{figure}[htbp] 
\vspace{-0.2cm}
\setlength{\abovecaptionskip}{0.1cm}  
\centering
\subfigure[]{
\includegraphics[scale = 0.28]{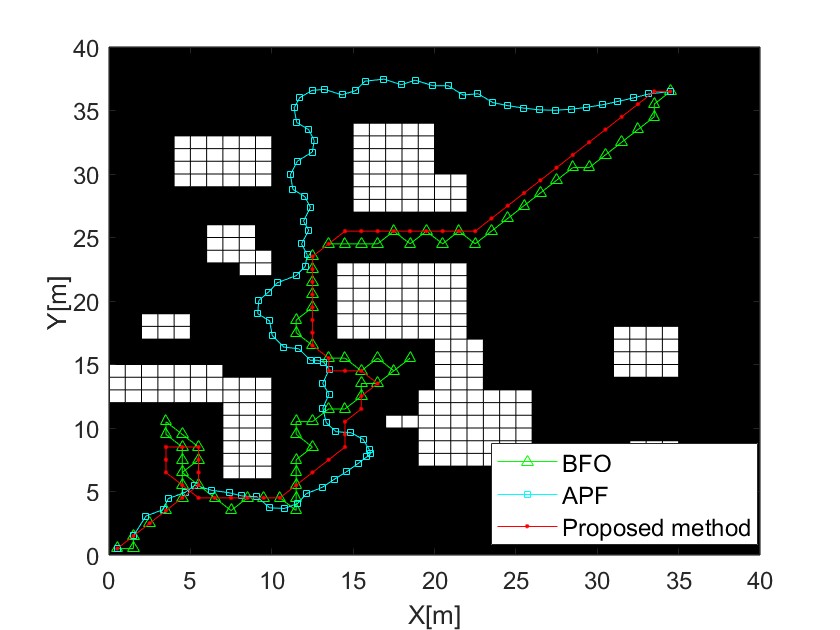}  
}\vspace{-4mm}  
\subfigure[]{
\includegraphics[scale = 0.4]{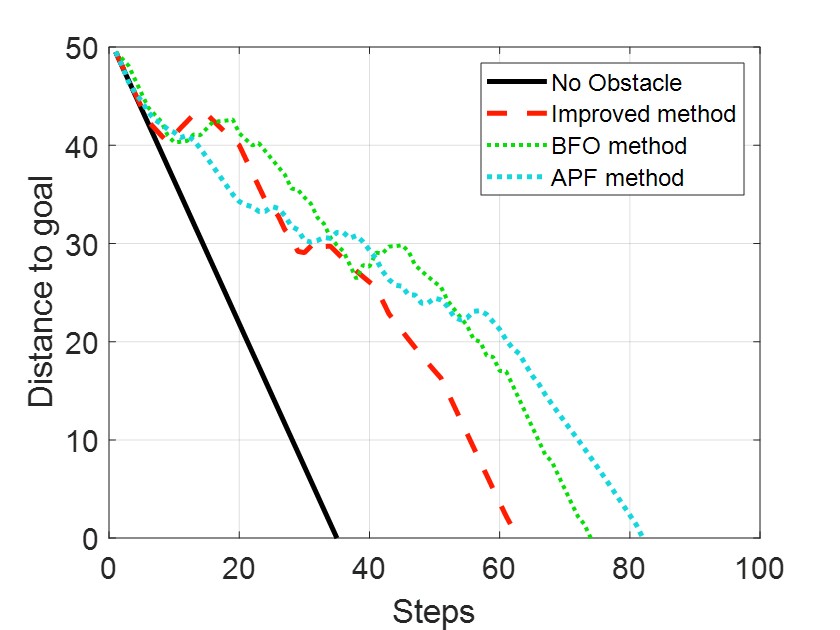}  
}
\caption{  Path trajectory and algorithm convergence }
\end{figure}

Figure 5(a) shows the paths generated by different algorithms, including the mainstream APF and BFO obstacle avoidance algorithms. By comparing the three paths, it is possible to see that the improved obstacle avoidance algorithm generates a solution with fewer fluctuations and shorter lengths.

For the given scenario, by computing the real-time distance between the robot and the goal position to compare the convergence characteristics of the three algorithms, as shown in figure 5(b). Considering path length, smoothness, and convergence properties. It can be concluded that the path generated by the proposed algorithm is not only globally convergent but also locally optimal.

\subsubsection{Simulation 2 - tune parameters on the evaluation results}\label{s4.1.2}
The cost function and the improved ACO algorithm both contain different weight parameters as described in section 3 in this paper. In this part, the effects of different azimuth factors and smoothing factor $\zeta$ on robot real-time path are mainly discussed. The distance factor ensures that the sub-goal is globally convergent, while the azimuth factors are responsible for the smoothness of the global path. To compares the effects of different weight parameters on the simulation results. Three sets of parameters are selected for simulation comparison. Finally, a set of optimal weight parameters is determined. Table 1 lists three different sets of parameters.

\begin{table}[ht]
\renewcommand\tabcolsep{4.0pt}  
\caption{Different parameters} \label{t1}
\centering
\begin{tabular}{p{1.7cm}p{0.4cm}p{0.4cm}p{0.4cm}p{0.4cm}p{0.4cm}}
\hline
\centering
\textbf{groups} &    &  \textbf{Parameters}  &    &    \\ \cline{2-6}     
\centering
      &$\alpha$  & $\beta$      & $\omega$ & $\delta$ & $\zeta$ \\ \hline  
\centering
First group   & 4     & 1.8     & 1        & 1      & 0    \\
\centering
Second group  & 4     & 1.8     & 1        & 0.7    & 0.3   \\
\centering
Third group   & 4     & 2       & 3        & 0.7    & 0.3   \\ \hline    
\end{tabular}
\end{table}

As shown in figure 6, the simulation program generates paths based on three different groups of parameters. It should be noted that this path is only one of the results of the simulation program that runs multiple times. Figure 6 shows that different parameters have a significant influence on the path length and number of turns. 

\begin{figure}[htbp] 
\vspace{-0.2cm} 
\setlength{\abovecaptionskip}{0.1cm}  
\includegraphics[scale = 0.28]{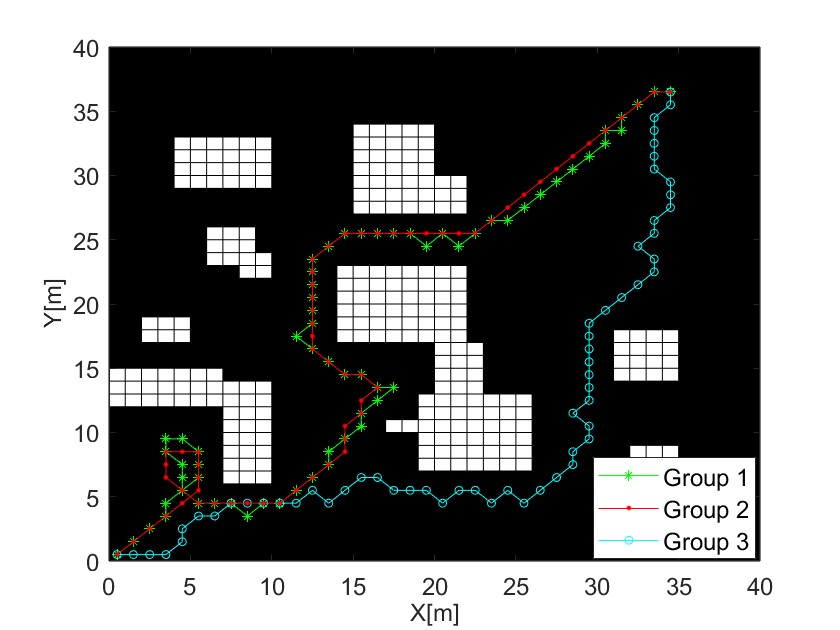}
\caption{ The effect of parameters on the simulation path}\label{f6}
\end{figure}

The differences between the first group and the second group of parameters are that the first group does not take into account the influence of the number of turns in the sub-path planning. While, the second group of parameters incorporates the turning factor, with the appropriate parameters, the result is smoother than the other path. In the third group, the azimuth factors are used to increase the smoothness of the path, but if its weight is too large, it will affect the convergence of the algorithm, as shown in figure 6.

In order to quantitatively analyze the different paths generated by the simulation, 10 simulations were performed for each set of parameters. Table 2 lists the path length and the number of turns of the global path generated by the simulation using different parameters, respectively. The simulation results consist of the best, worst, and average value.
\begin{table}[ht]
\caption{Simulation results in terms of different parameters} \label{t2}
\renewcommand\tabcolsep{4.0pt}  
\begin{tabular}{ccccccc}
\hline
\textbf{Results} & \multicolumn{2}{c}{\textbf{Group 1}}  & \multicolumn{2}{c}{\textbf{Group 2}}  & \multicolumn{2}{c}{\textbf{Group 3}} \\ \cline{2-7} 
\multicolumn{1}{c}{} & \multicolumn{1}{c}{\begin{tabular}[c]{@{}c@{}}Path\\ length\end{tabular}} & \multicolumn{1}{c}{corners} & \multicolumn{1}{c}{\begin{tabular}[c]{@{}c@{}}Path\\ length\end{tabular}} & \multicolumn{1}{c}{corners} & \multicolumn{1}{c}{\begin{tabular}[c]{@{}c@{}}Path\\ length\end{tabular}} & \multicolumn{1}{c}{corners} \\ \hline
Best    & 78.84  & 28      & 73.8    & 22   & 85.14   & 29 \\
Worst   & 88.68  & 35      & 79.25   & 27   & 98.91   & 51 \\ 
Average & 81.2   & 35      & 75.8    & 25   & 89.84   & 41 \\ \hline
\end{tabular}
\end{table}
With appropriate weight parameters, the simulation results have demonstrated that the proposed algorithm is capable of avoiding obstacles successfully and generating paths with a shorter length and few fluctuations.

\subsection{Robot experiments}\label{s4.2}
Refer to the above simulation results, the robot experiment platform of this paper is shown in figure 7(a). The proposed algorithm needs real-time information and process data online, including robot state, real-time environment information. So a $360^{\circ}$ LiDAR and an onboard computer installed on the EAI robot. EAI is a differential drive mobile robot with a maximum speed of 0.8 m/s and the drive wheel diameter of 125 mm, and this differential drive allows the implementation and measurements of path-planning algorithms. The architecture of the robot platform is illustrated in detail in figure 7(b). According to the actual experiment environment, the installation height of the LiDAR in two experiments is 0.6m. Also, the LiDAR scans the environmental information around the robot in real-time at 5Hz and publishes it. The onboard computer runs complex algorithms at high frequency, including LiDAR data collection and processing, obstacle analysis, and online path planning.
\begin{figure}[htbp] 
\vspace{-0.2cm} 
\setlength{\abovecaptionskip}{0.1cm}  
\centering
\subfigure[]{
\includegraphics[width=5cm]{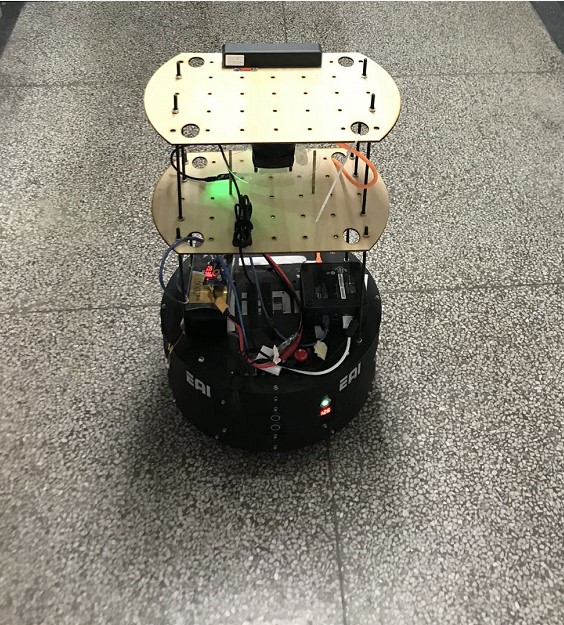}
}\vspace{-3mm}  
\subfigure[]{
\includegraphics[scale = 0.35]{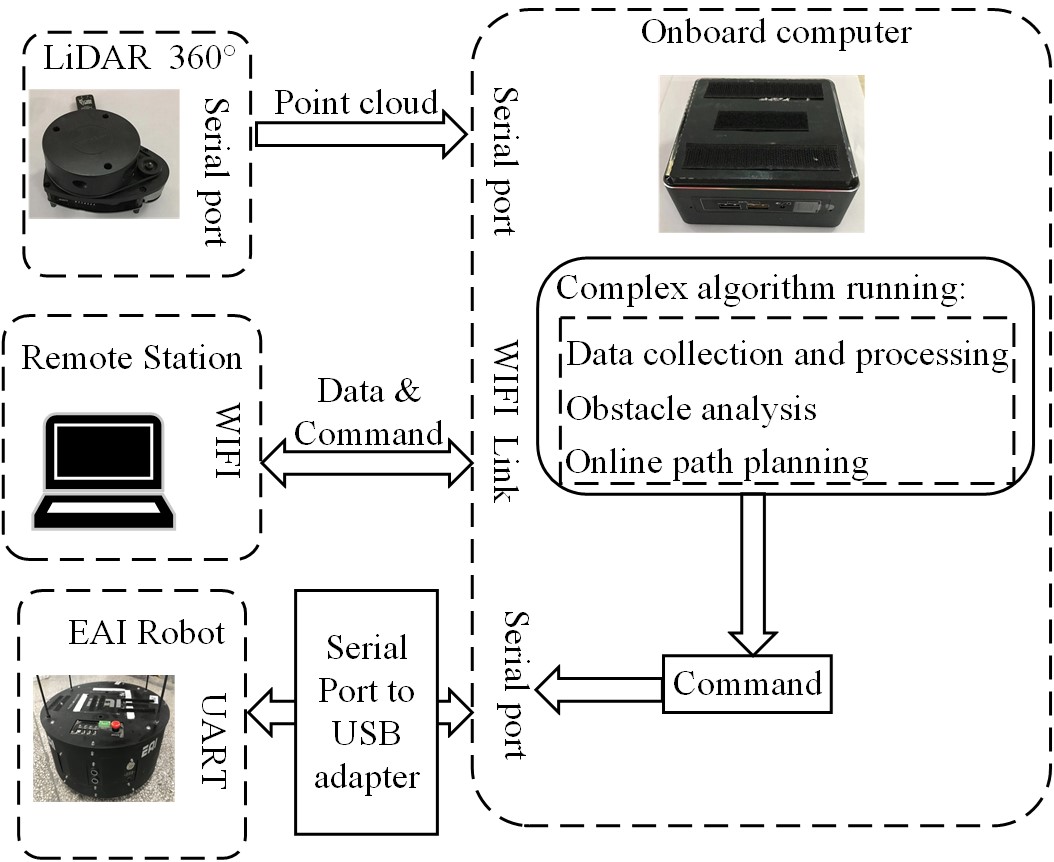}
}
\caption{  Mobile robot and block diagram of the architecture }
\end{figure}

In order to highlight the accuracy and practicality of the proposed algorithm to solve unexpected threats or uncertain obstacles, the first experiment scene was set in a real narrow corridor of 2.1m wide with static obstacles consisting of multiple cubes of $0.6 \times 0.6 \times 0.9 \mathrm{m}$. EAI robot radius is 0.203m and a safe distance of 0.2m from the obstacle. Further, the detection distance of the LiDAR is set to 1.2m to reduce the response time of the robot in scene 1. Scenario 2 is selected to verify the availability of the proposed algorithm in an unstructured, dynamic environment. Two real experimental scenarios with different effects are shown in figure 8. 
\begin{figure}[htbp] 
\vspace{-0.2cm} 
\setlength{\abovecaptionskip}{0.1cm}  
\centering
\subfigure[]{
\includegraphics[angle = 90,scale=0.87]{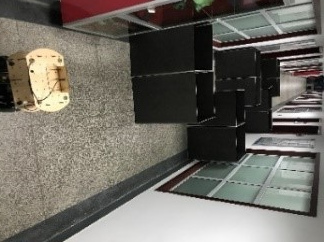} 
}\hspace{-1mm}
\subfigure[]{
\includegraphics[scale = 0.85]{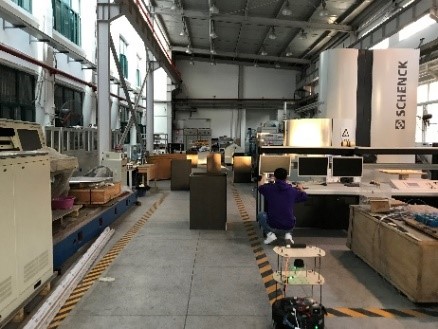} 
}
\caption{  Arrangement of real robot experiment environment }
\end{figure}
The onboard computer records the trajectory data of the robot, the real experiment environment is displayed in a certain proportion in MATLAB, and the saved robot trajectory is drawn at the same time. The complex U-shaped obstacles and L-shaped obstacles are used to verify the robustness and practicality of the proposed algorithm in figure 8(a), the closest distance between obstacles is about 0.9m, and the maximum width is 1.2m. Due to the extremely narrow corridor, the robot speed is 0.2m/s. In fact, the robot analyzes the LiDAR data in real-time to make the best decision in line with the current state. Similarly, the implementation of the algorithm is presented in the form of multiple consecutive instants in figure 9.

\begin{figure}[htbp] 
\setlength{\abovecaptionskip}{0.1cm}  
\centering
\subfigure{
\includegraphics[angle = 270,scale=1]{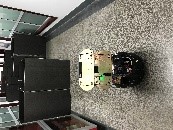}
}\hspace{-1mm}
\subfigure{
\includegraphics[angle=270,scale = 1]{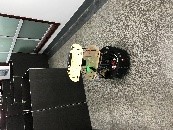}
}\hspace{-1mm}
\subfigure{
\includegraphics[angle=270,scale = 1]{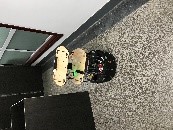}
}\hspace{-1mm}
\subfigure{
\includegraphics[angle=270,scale = 1]{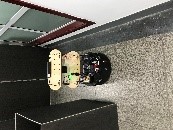}
}\vspace{-2mm}  
\subfigure{
\includegraphics[angle=270,scale = 1]{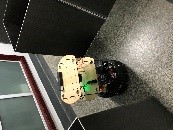}
}\hspace{-1mm}
\subfigure{
\includegraphics[angle=270,scale = 1]{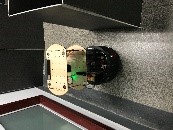}
}\hspace{-1mm}
\subfigure{
\includegraphics[angle=270,scale = 1]{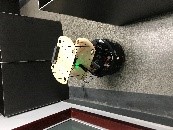}
}\hspace{-1mm}
\subfigure{
\includegraphics[angle=270,scale = 1]{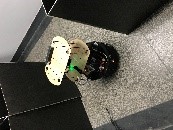}
}\vspace{-2mm}  

\subfigure{
\includegraphics[angle=270,scale = 1]{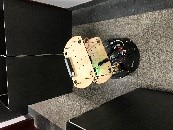}
}\hspace{-1mm}
\subfigure{
\includegraphics[angle=270,scale = 1]{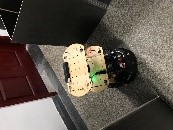}
}\hspace{-1mm}
\subfigure{
\includegraphics[angle=270,scale = 1]{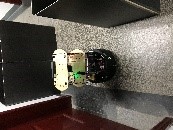}
}\hspace{-1mm}
\subfigure{
\includegraphics[angle=270,scale = 1]{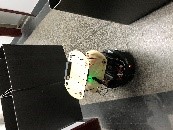}
}\vspace{-2mm}  
\subfigure{
\includegraphics[angle=270,scale = 1]{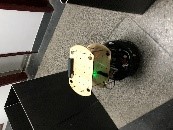}
}\hspace{-1mm}
\subfigure{
\includegraphics[angle=270,scale = 1]{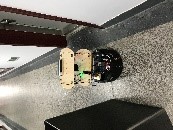}
}\hspace{-1mm}
\subfigure{
\includegraphics[angle=270,scale = 1]{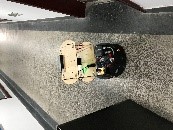}
}\hspace{-1mm}
\subfigure{
\includegraphics[angle=270,scale = 1]{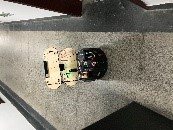}
}
\caption{Consecutive instants in the real experiment 1  }
\end{figure}

In the real experiment, a camera is mounted on the top of the robot and records the motion of the robot in video form. The robot successfully avoids the obstacle and reaches the goal position with minimal cost and smooth route, as shown in figure 10. However, the other methods fail to work well with LiDAR data, and the path length is long and not smooth enough.

\begin{figure*}[htbp] 
\hspace{-0.2cm}
\setlength{\abovecaptionskip}{0.1cm}  
\includegraphics[scale = 0.4]{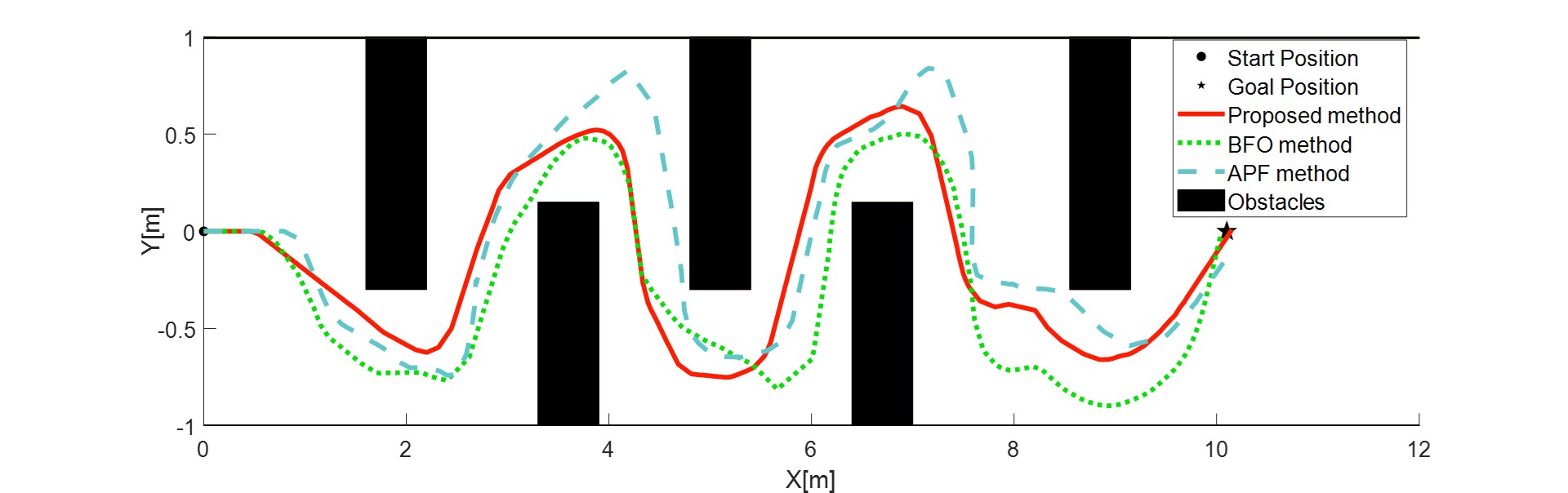}
\caption{Robot real trajectory in experiment 1}\label{f10}
\end{figure*}
In order to analyze the experiment results quantitatively, table 3 lists the results of each algorithm running five times at a speed of 0.2m/s. The recorded video can be downloaded from the link in Appendix A.

\begin{table}[ht]  
\caption{Comparison of experiment results of different algorithms} \label{t3}
\renewcommand\tabcolsep{4.0pt}  
\begin{tabular}{cccc}
\hline
\multicolumn{1}{c}{\textbf{Algorithm}}       & \multicolumn{1}{c}{\textbf{\begin{tabular}[c]{@{}c@{}}Optimal\\ Path length /m\end{tabular}}} & \multicolumn{1}{c}{\textbf{\begin{tabular}[c]{@{}c@{}}Average \\ Path length /m\end{tabular}}} & \multicolumn{1}{c}{\textbf{\begin{tabular}[c]{@{}c@{}}Average\\ Time /s\end{tabular}}} \\ \hline

Proposed method   & 15.5         & 16.2         & 81    \\
\begin{tabular}[c]{@{}c@{}}Conventional\\ ACO\end{tabular} & 16     & 17.06     & 85.3 \\
APF               & 18.2         & 18.5         & 92.5      \\ 
BFO          & 16.6         & 17.6       & 88        \\ \hline
\end{tabular}
\end{table}

It can be seen from table 3 that the method proposed in this paper takes less time, which means that the convergence speed is faster. Also, the smooth characteristics of the path have been compared in Table 2. Thus, considering time, path length, smoothness, and convergence properties, that the path generated by the proposed algorithm is not only globally convergent but also locally optimal.

As shown in figure 8(b), experiment 2 aims to verify the feasibility and robustness of the proposed algorithm under an unstructured, dynamic scenario with sudden external disturbances. In real experiments, the robot moves at a speed of 0.4 m/s and 0.6 m/s without external interference. Then, when the robot moves again at a speed of 0.6 m/s, a plurality of unexpected moving obstacles appears around the robot to hinder its movement. The trajectory of the robot is converted to convergence to illustrate the robustness of the algorithm, as shown in figure 11. At the same time, the corresponding video can be downloaded from the link in Appendix B.
\begin{figure}[htbp] 
\hspace{-0.2cm}
\setlength{\abovecaptionskip}{0.1cm}  
\includegraphics[scale = 0.4]{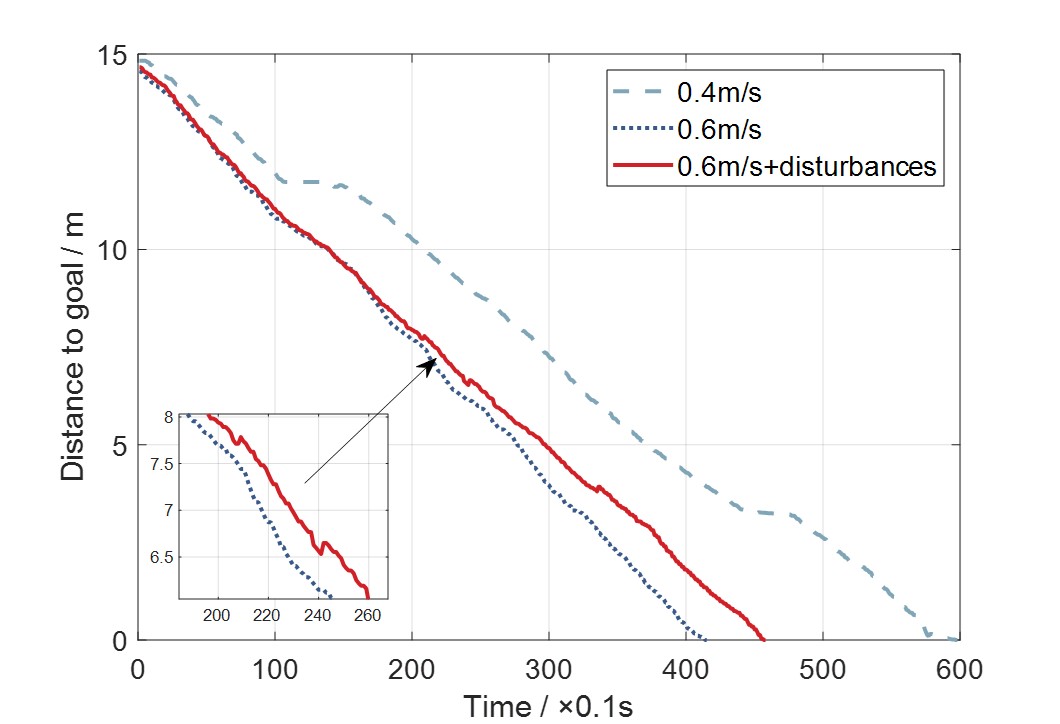}
\caption{Algorithm convergence in experiment 2}\label{f11}
\end{figure}

In Figure 11, in the absence of external disturbance, the convergence curve is relatively smooth, corresponding to the trajectory generated by the robot moving at a speed of 0.4 m/s and 0.6 m/s, respectively. For external disturbances, although there is jitter in the convergence curve due to sudden external disturbances, the jitter is minimal, which further demonstrate that our method can effectively deal with dynamic obstacles under an unstructured, changing scenario. 

\section{Conclusions }\label{s5}
In this paper, a new multi-constraints obstacle avoidance method using LiDAR was proposed. The main contributions of this method are its cost function for optimization is able to utilize the real-time LiDAR scanning data and the latest state estimation of the robot more comprehensively in dynamic environments. Compared with the traditional method, the advantage is that since the sub-goals and sub-paths are only related to the local regions, the optimal path is achieved at a lower cost in terms of computation and storage, thereby avoiding the construction of a complex global environment model and reducing the complexity.

The real experiments and simulation results also very strongly validated the feasibility and advantages of our method in contrast to other state-of-art solutions for this problem. 

The future work related to this proposed method include incorporating more dynamic characteristics of obstacles, finding better ways to fine-tune parameters of the cost function, testing the method to a even more complex environment, and implementing the method on a drone to extend its performance to 3D space. Another potentially interesting research direction is to include the "unseen" area of environments into optimization cost function, by extracting features of the point cloud to better describe the obstacles and then predicting potential threats to the planned path of the robot, based on features of the obstacles.

\section{Appendix A}\label{Appendix A}
Supplementary video material for experiment 1 related to this article can be found online at \url{https://youtu.be/8hr3ltbr11I}

\section{Appendix B}\label{Appendix B}
moving obstacles video for experiment 2 related to this article can be found online at \url{https://youtu.be/MFt9mqylN8w}

\end{document}